\documentclass[letterpaper, 10 pt, conference]{ieeeconf}  

\IEEEoverridecommandlockouts                              

\usepackage{amsmath,amsfonts,bm}

\def\eqref#1{equation~\ref{#1}}

\def\1{\bm{1}}

\DeclareMathAlphabet{\mathsfit}{\encodingdefault}{\sfdefault}{m}{sl}
\SetMathAlphabet{\mathsfit}{bold}{\encodingdefault}{\sfdefault}{bx}{n}

\usepackage{hyperref}
\usepackage{url}
\usepackage{graphicx}
\usepackage{xcolor}
\usepackage{stfloats}
\usepackage{wrapfig}
\usepackage{multirow}
\usepackage{caption}
\usepackage{subcaption}
\usepackage{algorithm}
\usepackage{algpseudocode}
\usepackage{booktabs}

\newsavebox{\largestimage}

\usepackage[absolute]{textpos}
\usepackage{amsthm}
\newtheorem{definition}{Definition}[section]

\definecolor{new}{HTML}{000000} 

\title{\LARGE \bf
Redundancy-aware Action Spaces for Robot Learning
} 

\author{Pietro Mazzaglia*$^{1}$ \and Nicholas Backshall*$^{2}$ \and Xiao Ma$^{2}$ \and Stephen James$^{2}$
\thanks{*Equal contribution.}
\thanks{$^{1}$IDLab, Ghent University, Technologiepark-Zwijnaarde 126, 9052 Gent, Belgium | {\tt\small pietro.mazzaglia@ugent.be}}%
\thanks{$^{2}$Dyson Robot Learning Lab, London, United Kingdom |
    {\tt\small firstname.lastname@dyson.com}}%
}

\begin{document}

\begin{textblock}{13}(1.5,0.25)
\centering \noindent\footnotesize © 2024 IEEE.  Personal use of this material is permitted.  Permission from IEEE must be obtained for all other uses, in any current or future media, including reprinting/republishing this material for advertising or promotional purposes, creating new collective works, for resale or redistribution to servers or lists, or reuse of any copyrighted component of this work in other works.
\end{textblock}

\maketitle
\thispagestyle{empty}
\pagestyle{empty}

\begin{abstract}
Joint space and task space control are the two dominant action modes for controlling robot arms within the robot learning literature. Actions in joint space provide precise control over the robot's pose, but tend to suffer from inefficient training; actions in task space boast data-efficient training but sacrifice the ability to perform tasks in confined spaces due to limited control over the full joint configuration. This work analyses the criteria for designing action spaces for robot manipulation and introduces ER (\textbf{E}nd-effector \textbf{R}edundancy), a novel action space formulation that, by addressing the redundancies present in the manipulator, aims to combine the advantages of both joint and task spaces, offering fine-grained comprehensive control with overactuated robot arms whilst achieving highly efficient robot learning. We present two implementations of ER, ERAngle (ERA) and ERJoint (ERJ), and we show that ERJ in particular demonstrates superior performance across multiple settings, especially when precise control over the robot configuration is required. We validate our results both in simulated and real robotic environments.
\end{abstract}
\noindent \textbf{Project website:} 
\texttt{\href{https://redundancy-actions.github.io}{redundancy-actions.github.io}}





\begin{figure}[t]
    \centering
      \savebox{\largestimage}{\includegraphics[width=.25\linewidth]{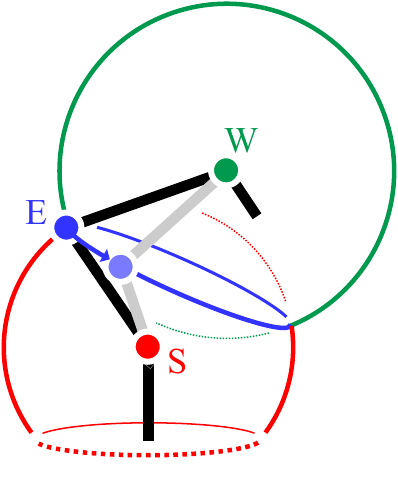}}%
  
  \hfill
   \begin{subfigure}[t]{0.45\textwidth}
    \centering
      \includegraphics[width=0.45\textwidth]{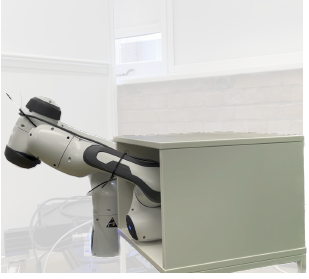}
      \hfill
      \includegraphics[width=0.45\textwidth]{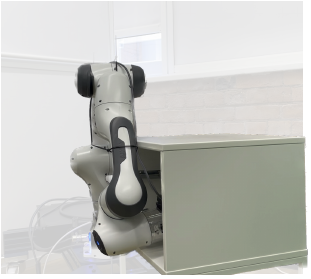}
    \caption{Manipulation in confined space}
    \label{fig:constrained}
  \end{subfigure}
  \begin{subfigure}[t]{0.45\textwidth}
    \centering
      \includegraphics[width=1.05\textwidth]{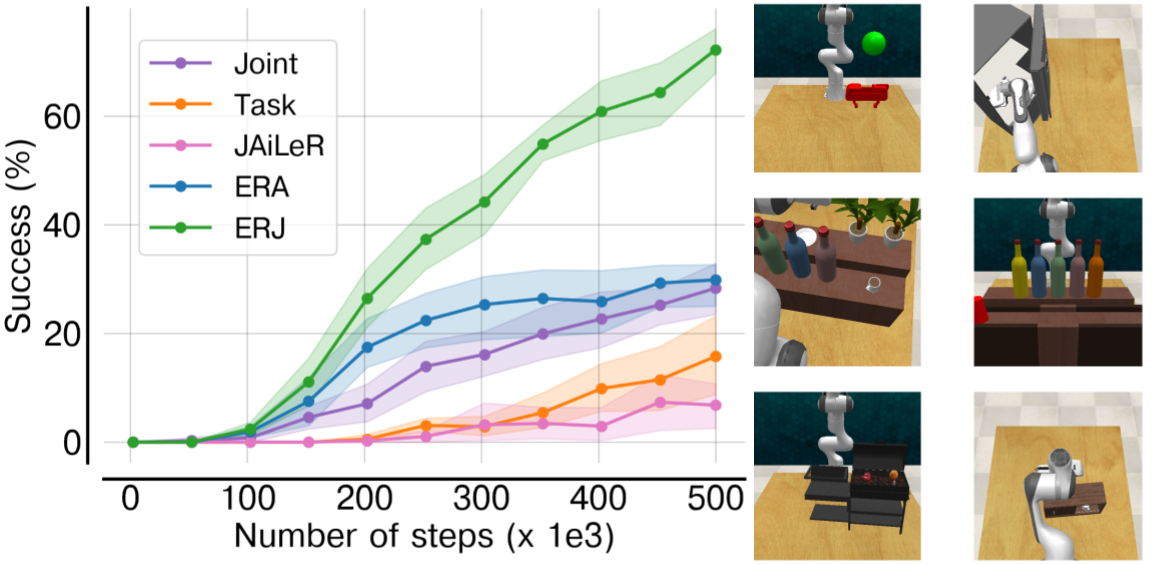}
    \caption{Performance on tasks requiring full-body control}
    \label{fig:performance_comparison}
  \end{subfigure}

\caption{\textbf{(a)} The configuration of the entire arm is crucial for manipulation in confined spaces, e.g., removing an object from the cabinet. A correct elbow position enables the robot to enter the cabinet (left), but an incorrect elbow position causes unsuccessful execution of the action (right). \textcolor{new}{\textbf{(b)} Our ER action spaces (ERJ and ERA) outperform other action spaces in tasks where control over the entire robot configuration is required (6 tasks shown on the right).}} 
\end{figure}

\section{INTRODUCTION}
\label{sec:intro}

The choice of the action representation, in robot learning,  greatly affects task learning efficiency and precision of robot control, as it defines the space of learnable actions for the agent. The two predominant action representations in the current landscape of robot learning research are joint space coordinates and \textcolor{new}{task space coordinates (also known as end-effector coordinates)}. Their popularity is due to the minimal computation overhead required and their implementation simplicity on most robotic platforms, even in the absence of joint-torque sensors. 


Actions in joint space provide joint-level control over the entire configuration of the robot but typically lead to less sample-efficient behaviour learning. Task space coordinates simplify the learning process, with actions being in the same space as the task, but cannot provide full-body control in overactuated arm configurations. For example, as shown in Figure \ref{fig:constrained}, using task space control to stretch deep into a cabinet is difficult due to the free motion in the elbow.

To address the shortcomings of these two action spaces, we propose a new flexible and efficient action space formulation for overactuated arms: \textbf{E}nd-effector \textbf{R}edundancy (\textbf{ER}). This formulation allows actions to be provided in task space coordinates, facilitating behaviour learning, while enabling control over the full robot's configuration, by addressing any redundancies in the robot's morphology. The ER action spaces can be parameterised and controlled in various ways, and in this paper, we analyse two concrete realisations of it: \textbf{ERA}ngle (\textbf{ERA}) and \textbf{ERJ}oint (\textbf{ERJ}). 

Across a series of quantitative reinforcement learning (RL) experiments, we demonstrate that, when the task requires precise control, the ER action spaces outperform classical action spaces, such as joint space and task space, \textcolor{new}{and hybrid learned action spaces, such as JAiLeR \cite{Kumar2021Jailer}}, both in efficiency and performance, as shown in Figure \ref{fig:performance_comparison}.

\textbf{Contributions.} Our contributions can be summarised as:
\begin{itemize}
    \item a new class of action spaces, ER, enabling full-body control of the robot arm's pose, while retaining efficient learning performance -- and we describe and implement two concrete realisations of it: ERA and ERJ;
    \item an extensive RL-based evaluation of our approach in simulation with RLBench \cite{James2019RLBench, Rohmer2013VREP, james2019pyrep}, showing the change in action space is crucial in tasks requiring precise control over the arm configuration, without compromising performance in other tasks;
    \item additional studies on ERJ demonstrate its applicability for real-world imitation learning, showing success in tasks where the task space fails, and the possibility to work on robots having complex morphologies with higher degrees of freedom.  
\end{itemize}
We release the code for our experiments, including the implementations of ERJ and ERA and the assets for the RLBench tasks requiring full-body control,  to facilitate future research about redundancy-aware action spaces for manipulation.

\section{RELATED WORK}
\label{sec:background}

\textbf{Action spaces for robotic manipulation.} When controlling a robot arm, commands need to be sent to the controller in configuration space, e.g. joint positions or torques. However, it is possible to operate in Cartesian task space coordinates, by using techniques such as inverse kinematics (IK) solvers or operational space controllers (OSC) \cite{Nakanishi2008OSC} that translate Cartesian commands in the configuration space. Previous work tends to associate working in task space with higher performance \cite{matas2018sim, Varin2019ActionSpaceManipulation, martin2019variable, Zhu2020Robosuite}. However, the tasks analysed generally require only partial control over the robot joint configuration. 
Recent work has attempted to link joint control to task space coordinates, either learning a joint space controller that can drive the agent in task space \cite{Kumar2021Jailer} or using the forward kinematics to translate joint coordinates into task space coordinates and learning multi-action space policies \cite{Ganapathi2022IKP}. While joint and task space controllers are both common for low-level fine-grained control, other work has focused on the problem of solving long-term tasks or low data efficiency adopting high-level action primitives for control, such as pushing and grasping \cite{Zeng2018PushGrasp}, sliding and turning \cite{Dalal2021RAPS, Nasiriany2022MAPLE}, 3D visual coordinates in the workspace \cite{James2022ARM, Mazzaglia2024IDA}, or a hierarchical combination of task-space controllers \cite{Sharma2020HierarchicalControllers}. 
\textit{Our work} proposes a set of action spaces that, similarly to task and joint spaces, works at the lowest level of the control hierarchy, so that future work could easily build on top of ours. We deliberately choose to focus on IK-based solvers for translating from task space to configuration space, as they are the most accessible across all robot configurations. 

\textbf{Robot learning.} Learning paradigms for robotics, such as reinforcement learning \cite{Fujimoto2018TD3, Haarnoja2019SAC} and imitation learning \cite{Ho2016GAIL, Florence2021ImplicitBC}, or a combination of both \cite{Rajeswaran2018DAPG, Zhan2020FERM}, can help automate behaviour learning in robotics. While previous work has shown the potential of robot learning when training from large chunks of data \cite{Kalashnikov2018QTOPT, Openai2019Rubik, Lu2021AWOPT}, recent work has proven that robot learning can lead to data-efficient learning too \cite{Smith2022WalkPark, Zhao2023ACT}. There are several challenges that are specific to robot learning, such as extracting useful features from high-dimensional observations \cite{radosavovic2023real, Nair2022R3M}, choosing adequate policy parameterisations \cite{seyde2021bang, james2022bingham}, and online adaptation \cite{Kumar2021RMA}. \textit{In this work}, we focus on the problem of finding an optimal action space, that allows both precise control and efficient learning.

\textbf{Redundant manipulators control.} Despite 6 DoF robot arms being sufficient for many tasks and industrial settings, the redundancy in overactuated arms ($>6$ DoF) can be useful for manipulability \cite{Yoshikawa1985Manipulability}, torque optimisation \cite{Suh1987Torque}, obstacle avoidance  \cite{Nakamura1987Obstacles} and singularity robustness \cite{Chiaverini1997Singularity, Cheng1998Resolution}. Redundancy resolution is a long-standing problem in robotics \cite{Hollerbach1987RedundancyTorque, hsu1989dynamic,Seraji1989} whose solutions often require taking into account the structure of the robot, by defining additional constraints \cite{Howard2009Constrained, Lin2015NullSpace}, e.g. an arm angle \cite{Shimizu2008ArmAngle}, and/or analytical solutions \cite{He2021Franka}, or using RMPs \cite{Ratliff2018RMPs} where the redundancy is resolved using a default joint configuration, but not controlled.
\textit{Our work} aims to apply classical redundancy control methods to robot learning. 

\section{ANALYSIS OF CURRENT ACTION SPACES}
\label{sec:deep_dive_into_action_modes}

In this section, we explore the two predominant control methods within the current landscape of robot learning research: task space and joint space coordinates.

\subsection{Task Space}

\begin{definition}[Task Space]
     An action space $\mathcal{A}_{ee}$ with 6 DoF, consisting of $\mathcal{A}_T\in \mathbb{R}^3$ for translations, and an over-parameterised quaternion action $\mathcal{A}_{R}\in \mathbb{R}^4$ for rotations, of the end-effector (ee).
\end{definition}
The task space directly maps to the task objective: manipulating the end-effector of the robot for environmental interaction. This alignment has been shown to improve sample efficiency by simplifying the optimisation process~\cite{matas2018sim, plappert2018multi}.
However, it falls short of providing comprehensive control for overactuated robot arms. For example, the redundancy of overactuation arises when using the 6 DoF task space to control a robot arm with $>6$ DoF, where multiple joint configurations may achieve the same desired end-effector pose. As a result, the joint configuration space remains under-controlled and the redundant joint(s) can vary whilst maintaining the end-effector pose, creating free-motions. This limitation becomes evident in scenarios where precise control of the entire robot is imperative for successful task execution, such as obstacle avoidance or reaching into confined spaces like cramped cupboards, as shown in Figure~\ref{fig:constrained} with a 7 DoF arm. In these situations, task-space actions lead to collisions and sub-optimal outcomes.

\subsection{Joint Space}
\begin{definition}[Joint Space]
    An action space $\mathcal{A}_J \in \mathbb{R}^n$ where $n$ is the number of joints in the robot arm.
\end{definition}
Joint space actions are adept at collision avoidance through complete joint-level control of the robot's morphology. With access to all joints, this action space enables the robot to accurately reach any configuration and address the issue discussed in Figure~\ref{fig:constrained}. However, the $\mathcal{A}_J$ joint space is not aligned with the Cartesian task space the robot is interacting with. Mapping from $\mathcal{A}_J$ to this task space requires the policy to understand the underlying non-linear kinematics of the robot, inherently increasing the complexity of the task. As a result, the joint space tends to have worse sample efficiency than the task space.

\subsection{Pivotal Criteria of Action Spaces for Robot Learning}
The trade-off between controllability and efficiency leads to the question: \textit{does an ideal action space which strikes a balance between these aspects exist?}

To answer this question, we define three pivotal criteria of action spaces for robot learning to be satisfied for achieving comprehensive and efficient control:
\begin{itemize}
    \item \textit{\textbf{Alignment}}: the action space should be aligned with the 6 DoF task space, to enhance sample efficiency in robotic manipulation and simplifying grasp locations and movement trajectories. 
    \item \textit{\textbf{Discriminability}:} \textcolor{new}{control} over the robot configuration is expected to be consistent for identical state-action pairs and to lead to unique solutions, i.e. no free-motion.  
    \item \textit{\textbf{Validity}}: the agent's desired actions should result in achievable configurations which obey the kinematics constraints and joint limits of the robot. 
\end{itemize}

The task space naturally satisfies the alignment criteria; however, it lacks validity when poses violate kinematic constraints, and discriminability when controlling overactuated robot arms (due to free-motion). The joint space, on the other hand, provides both validity and discriminability, but fails to meet the criteria for alignment.

\section{ER Action Spaces for Robot Learning}
\label{sec:method_theory}

In this work, we introduce \textbf{E}nd-effector \textbf{R}edundancy (\textbf{ER}) action spaces, a new family of action spaces for robot learning with overactuated arms. ER action spaces combine the advantages of both task and joint spaces, providing accurate control of the entire arm and allowing the agent to efficiently operate in task space.

\subsection{ER Action Spaces}

\begin{definition}[ER Space]
     An action space of $\mathcal{A}_{er} = (\mathcal{A}_{ee}, \mathcal{A}_{r})$, where $\mathcal{A}_{r} \in \mathbb{R}^{n-6}$ are the redundancy constraints to the free-motion present in overactuated arms when using end-effector pose control (for robots with $n$ DoF).
\end{definition}

Given an $N$ DoF arm (where $N > 6$), the ER space builds upon the task space, adding $N - 6$ dimensions to the action, which are used to constrain the free motions.
Adopting appropriate parameterisations to address redundancy control problems provides control over free motions not available in the task space (\textit{discriminability}). By controlling the robot in a hybrid space, which contains both task space coordinates and additional controls for any redundancies, the agent can infer the relation between the task and the operational space more easily (\textit{alignment}).

Translating from task space to joint space coordinates is commonly completed using IK solvers or OSC controllers. Additionally, RMPs provide a similar utility \cite{Ratliff2018RMPs} whilst also providing ``redundancy resolution and damping"; however, this resolution uses a controller of the form:
$$\mathbf{f}_d(\mathbf{q}, \dot{\mathbf{q}}) = \alpha (\mathbf{q}_0 - \mathbf{q}) - \beta \mathbf{q}$$
\noindent which resolves the redundancy in joint space, using a constant default joint configuration ($\mathbf{q}_0$), typically set to 0. To control this free-space, $\mathbf{q}_0$ would have to be parameterised which equates to learning a task in joint space. 

\textcolor{new}{To maintain precise control over the entire robot configuration, we opt for using position-based control through IK solvers. Compared to OSC controllers, IK solvers for position control tend to transfer more easily to real robots \cite{Kumar2021Jailer} and do not require direct access to the motor torques. Furthermore, they allow fast computation of solutions when working in relative coordinate systems, such as the delta actions typically used in RL, using the Jacobian pseudo-inverse method. This allows our method to be agnostic to the system of coordinates used, either relative or absolute.}

By controlling the end-effector in Cartesian and parameterising redundancies, the ER space aims to combine the sample efficiency of working in task space with full control over the joint configuration. 
However, in order to lead to achievable configurations, the agent needs to output valid combinations of $ee$ and $redundancy$ commands. The \textit{validity} of the action space will depend on how the redundancy is parameterised. Two concrete realisations of the ER space are introduced in the rest of this section. 

\subsection{ERAngle - Arm Angle Control}

\begin{figure}[t]
    \centering
    \begin{subfigure}[b]{.49\columnwidth}
        \centering
        \includegraphics[height=.6\textwidth]{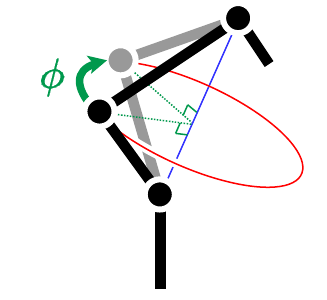}
        \caption{ERAngle (ERA)}
        \label{fig:e3a_diagram}
    \end{subfigure}
    \hfill
    \begin{subfigure}[b]{.49\columnwidth}
        \centering
        \includegraphics[height=.6\textwidth]{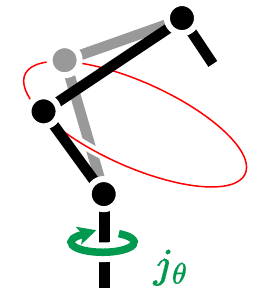}
        \caption{ERJoint (ERJ) }
        \label{fig:e3j_diagram}
    \end{subfigure}
    \caption{\textbf{ER family of action spaces.} (a) ERA is constrained directly by the angle of the elbow around the axis from the shoulder to the wrist ($\phi$). (b) ERJ is constrained through removing joints from the inverse kinematics and controlling these separately (the base joint $j_\theta$ in the diagram).}
    \label{fig:e3a_e3j}
\end{figure}

In classical control literature, one popular solution to the redundancy control problem, is to describe the redundancy through the adoption of an additional angle, often referred to as the arm angle \cite{Shimizu2008ArmAngle}. Applying this solution to robot learning, we define the \textbf{ERAngle} or \textbf{ERA} space as:
\begin{definition}[ER Angle space]
     An action space $\mathcal{A}_{era} = (\mathcal{A}_{ee}, \mathcal{A}_{\phi})$, where $\mathcal{A}_{\phi} \in \mathbb{R}^{n-6}$ are the rotation angles around the lines connecting the centre of spheres which create free-motion (for robots with $n$ DoF).
\end{definition}

For example, as shown in Figure~\ref{fig:e3a_diagram}, for a 7 DoF arm with a known robot kinematics model, given the gripper pose and rotation angle $\phi$, there exists a closed-form solution to the position of the elbow, $\mathbf{p_E \in \mathbb{R}^3}$
\begin{equation}
    \mathbf{p_E} = \mathbf{c_E} + r_E \cdot \left( \mathbf{t_E} \cdot \cos(\phi) + \mathbf{b_E} \cdot \sin(\phi) \right)
\end{equation}
where $\mathbf{c_E} \in \mathbb{R}^3$ and $r_E$ are the centre and radius of the elbow free-motion circle, and $\mathbf{b_E} \in \mathbb{R}^3$ and $\mathbf{t_E} \in \mathbb{R}^3$ are bi-tangent the tangent to this circle at the desired elbow angle $\phi$ (where $0^{\circ}$ is the highest point on the circle). 
We can now add $\mathbf{p_{E}}$ as an additional constraint to the IK solver and obtain the joint positions of the arm.

ERA successfully addresses the joint redundancy issue of arms with $>6$ DoF and satisfies the discriminability and alignment criteria. Nevertheless, with the additional IK constraint imposed on the elbow joint, the solution of the IK problem will face a more complex optimisation landscape, causing some actions of the action space to be invalid. 

\subsection{ERJoint - Joint-First-Control}

To address the problem of finding IK solutions with ERA, we further introduce \textbf{ERJoint} or \textbf{ERJ} space, which controls the redundancies by directly controlling some joints of the robot arm (applying familiar redundancy control methods \cite{He2021Franka} to robot learning). Formally, we define ERJ as:
\begin{definition}[ER Joint space]
 An action space $\mathcal{A}_{erj} = (\mathcal{A}_{ee}, \mathcal{J}_\theta)$, where $\mathcal{J}_\theta \in \mathbb{R}^{n-6}$ are the scalar actions that control the position of the chosen arm joints (for robots with $n$ DoF).
\end{definition}

Controlling additional joints can be seen as a way of reducing the number of optimising variables solved with the IK. Specifically, in the case of a 7 DoF arm, we set the chosen joint to the desired target position and resolve for the remaining $7-1=6$ DoF coordinates of the robot in task space by solving the IK. Ideally, ERJ should allow both flexible control of the robot, and a higher IK solver success rate, given that fewer constraints are imposed. However, this comes at the cost of establishing which joints should be controlled separately. 

\begin{figure*}[t]
    \centering
    \begin{minipage}[t]{.62\textwidth}
        \centering
        \includegraphics[width=\textwidth]{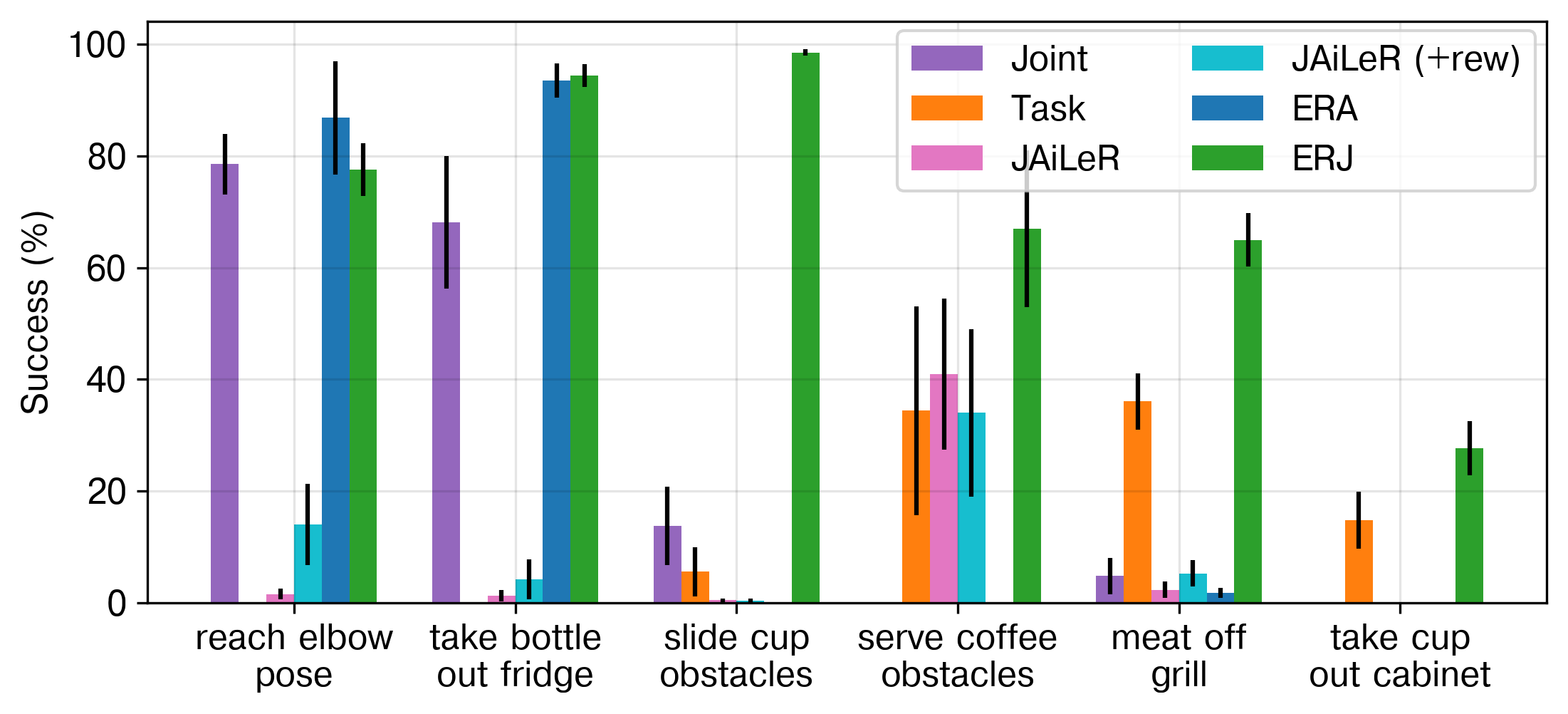}
        \caption{Results on 6 tasks where control over the entire robot arm configuration is required to succeed or perform reliably \textcolor{new}{(4 runs).}} %
        \label{fig:elbow_tasks}
    \end{minipage}%
    \hfill
    \begin{minipage}[t]{0.35\textwidth}
        \centering
        \includegraphics[width=\textwidth]{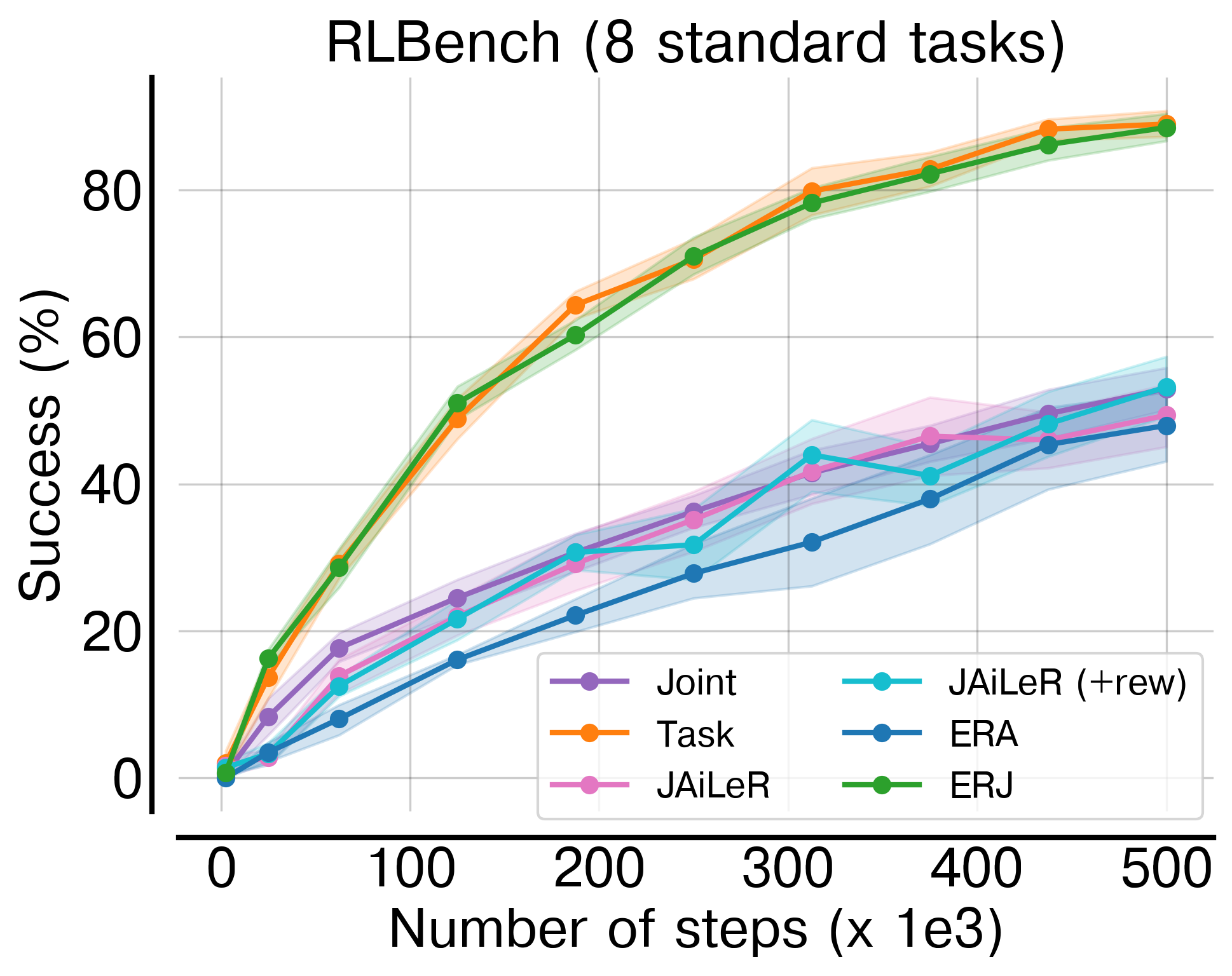}
        \caption{Comparing action modes on a standard suite of 8 manipulation tasks. \textcolor{new}{(4 runs)}.}  
        \label{fig:non_elbow_tasks}
    \end{minipage}
\end{figure*}

\textbf{Joint selection.}
The selection of the joints that are controlled separately is crucial, as this influences how the IK problem is solved for the other joints. Consider the forward kinematics of a 7 DoF robot as a chain. Intuitively, fixing the intermediate joints will break the chain. Although this does not introduce additional constraints to the optimisation problem of solving the IK, extra efforts are needed to customise IK solvers as the chaining property of the robot kinematics is broken. Thus, the first (base) or the last joint (wrist) are intuitively more practical choices. We choose to control the base joint by default but analyse how joint selection affects performance in the Experiments section. 

\section{Experiments}
\label{sec:experiments}

In our experiments, we aim to showcase the performance of the different implementations of the ER space, ERA and ERJ, in several tasks, compared to the common task and joint space action modes. \textcolor{new}{We also compare ER to a learning-based action space: JAiLeR \cite{Kumar2021Jailer}, which learns a controller through RL to translate goal end-effector coordinates into joint velocities for position control. When employing JAiLeR, a task policy is trained to provide goals in end-effector coordinates to the controller of JAiLeR, which is trained simultaneously to the policy, to maximise the control reward:}
$$r_\textrm{ctrl} = \textrm{exp}(-\lambda_\textrm{err}\Vert\delta x\Vert^2) - \lambda_\textrm{eff} \Vert \ddot{q} \Vert. $$
\noindent \textcolor{new}{Since JAiLeR was only tested for goal-reaching tasks, we introduce an additional baseline JAiLeR (+rew), where the controller reward is also rewarded with the task reward, weighted by a $\lambda_\textrm{task}=0.05$ coefficient ($\lambda_\textrm{err}=20$ and $\lambda_\textrm{eff}=0.005$, as for the original paper)}.

We present results both in simulation (where actions are delta positions) and real-world settings (where actions are absolute positions), using RL and imitation learning, respectively. 
Given the higher performance obtained using ERJ, compared to ERA, we extend our study of this method, analysing how joint selection affects ERJ and showing how the method easily scales and proves useful in a complex simulated robot with more than 7 DoFs. \textcolor{new}{In all Figures, we use lines or bars to represent mean statistics over multiple runs (indicated in captions) and shaded areas or black lines for representing the standard deviation, for single tasks, or confidence intervals, for plots aggregating multiple tasks, calculated according to \cite{Agarwal2021RLiable}.}


\subsection{Reinforcement Learning in Simulation}

We train RL agents in RLBench~\cite{James2019RLBench}, across two sets of tasks: (i) 6 `\textit{Full-body}' tasks which specifically require or benefit from full control of the robot joint configuration, and (ii) 8 `\textit{Standard RLBench}' tasks that are commonly employed for RL evaluation. In both settings, we provide dense reward functions for all tasks, to drive the agent's exploration towards success, and we adopt SAC \cite{Haarnoja2019SAC}, \textcolor{new}{taking as inputs two camera images, wrist and third-view, and a set of proprioceptive states}. \textcolor{new}{To calculate the robot joint configuration for IK-based methods, we rely on the internal IK solver of CoppeliaSim employing the pseudo-inverse Jacobian. We found it useful to adopt a damped least squares solver for ERA, in order to reduce the invalid actions rate, while this change had no significant effect for the other methods. On our machines, computation and execution times with ERA (13ms) and ERJ (19ms) are comparable with task space control (12ms).
}

\begin{figure*}[t]
\centering
\captionof{table}{We compare success rate (mean ± std err) of \textcolor{new}{Joint (oracle)}, Task and ERJ over 5 real-world manipulation tasks.}
\label{tbl:real_results}
\begin{tabular}[t]{lccccc}
\toprule
&Reach Cup&Remove Cup&Reach \& Remove Cup & Retrieve Bear & Press Button Elbow \\
\midrule
Joint (oracle)&\textbf{100}\%&\textbf{90}{\scriptsize±9.5}\%&\textbf{90}{\scriptsize±9.5}\%&\textbf{100}\%&\textbf{100}\%\\
Task&0\%&40{\scriptsize±15.5}\%&0\%&0\%&0\%\\
ERJ&\textbf{100}\%&\textbf{80}{\scriptsize±12.6}\%&\textbf{90}{\scriptsize±9.5}\%&\textbf{100}\%&\textbf{100}\%\\
\bottomrule \\
\end{tabular}
\begin{minipage}{\textwidth}
    \centering
    \begin{subfigure}[b]{.185\textwidth}
        \centering
        \includegraphics[trim=3.3cm 0 2cm 0,clip,width=.8\textwidth]{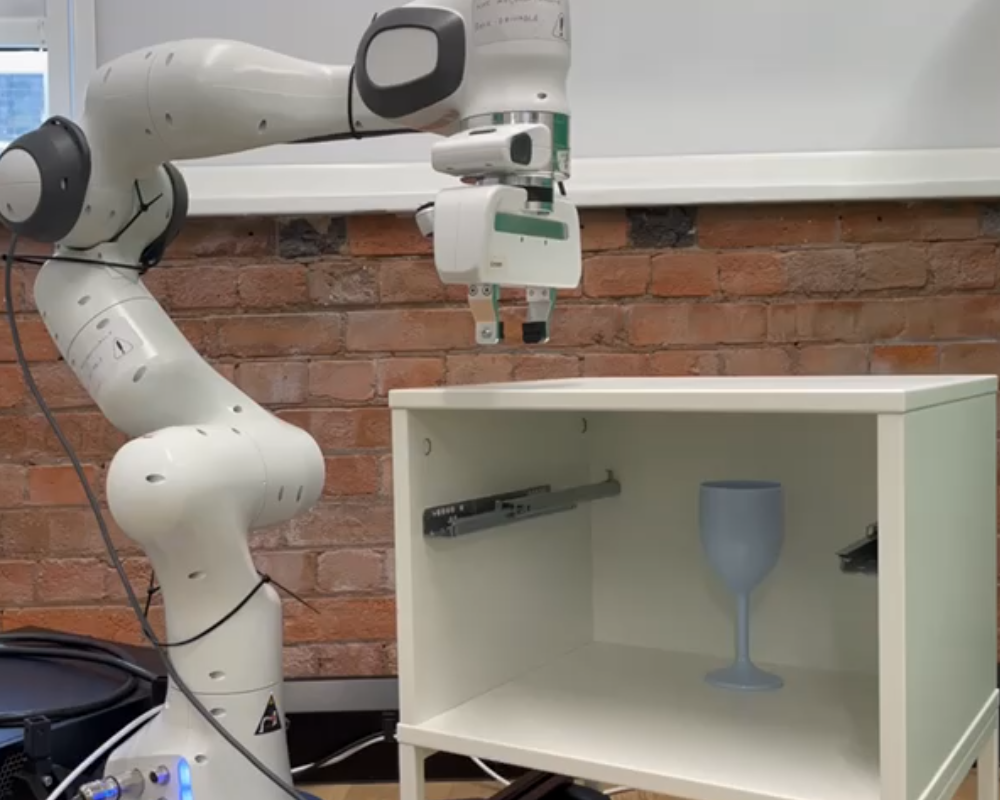}
        \caption{Reach Cup}
    \end{subfigure}
    \begin{subfigure}[b]{.185\textwidth}
        \centering
        \includegraphics[trim=3.3cm 0 2cm 0,clip,width=.8\textwidth]{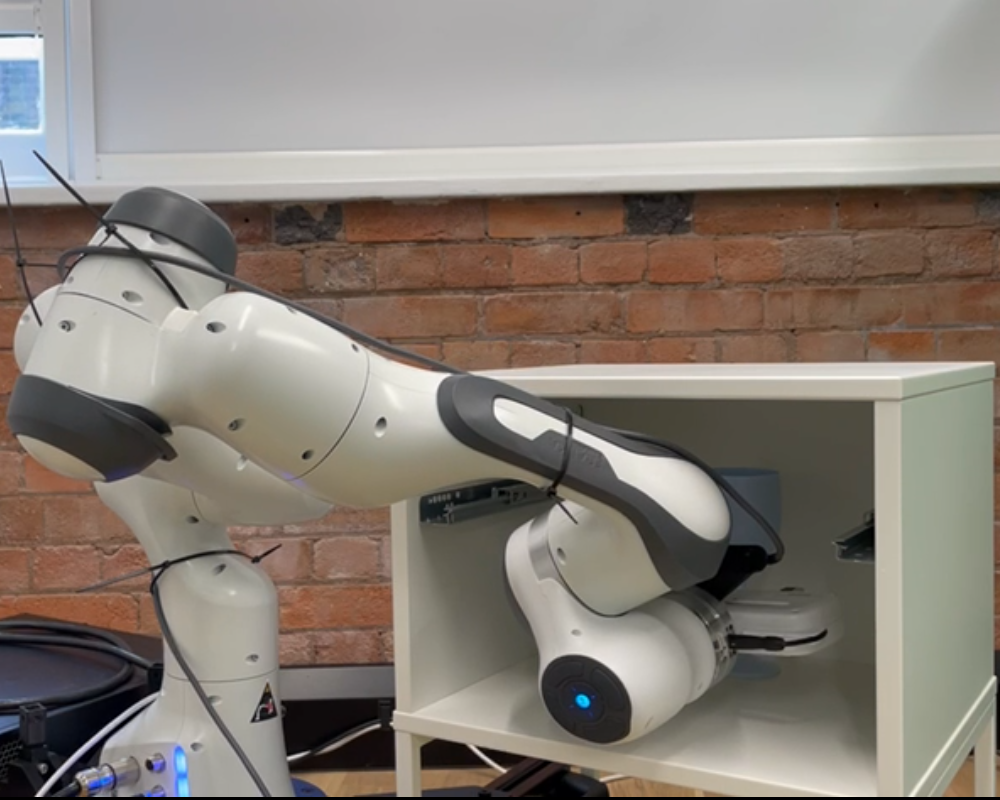}
        \caption{Remove Cup}
    \end{subfigure}
    \begin{subfigure}[b]{.185\textwidth}
        \centering
        \includegraphics[trim=5.3cm 0 0 0,clip,width=.8\textwidth]{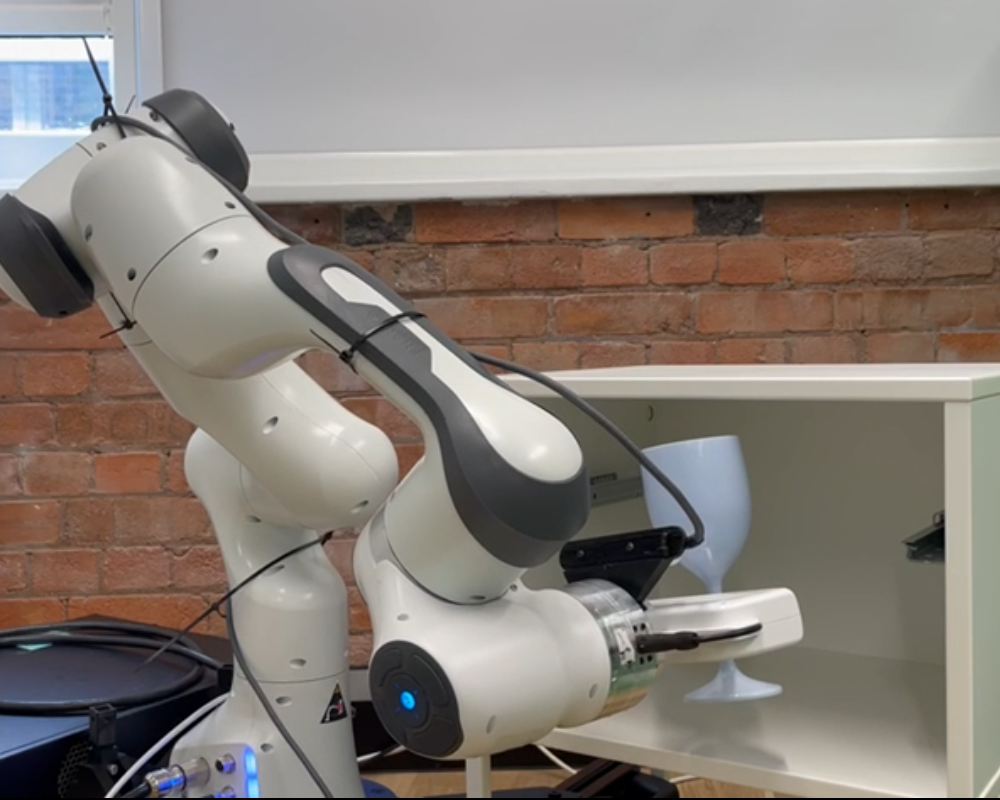}
        \caption{Reach \& Remove Cup}
    \end{subfigure}
    \begin{subfigure}[b]{.185\textwidth}
        \centering
        \includegraphics[width=.8\textwidth]{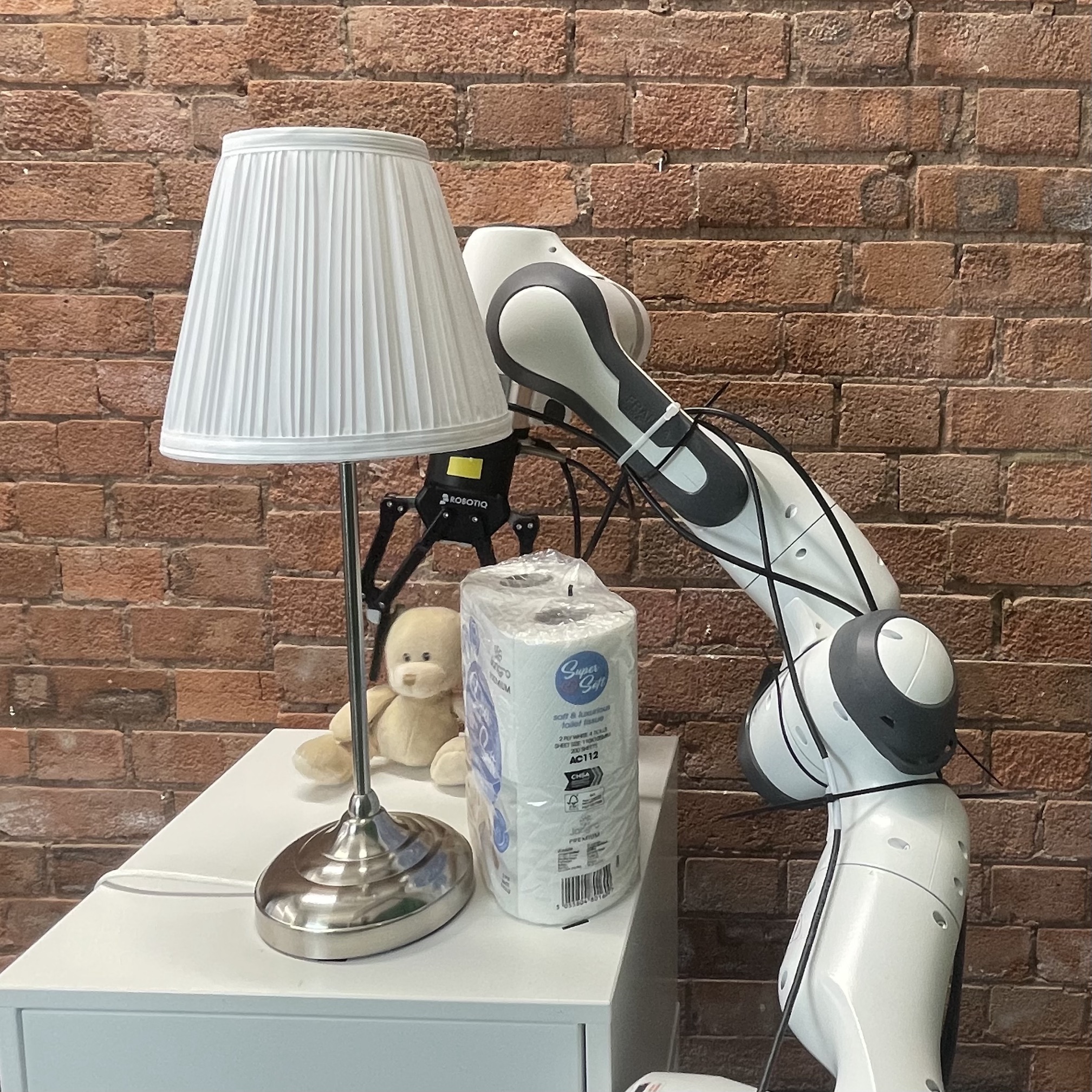}
        \caption{Retrieve Bear}
    \end{subfigure}
    \begin{subfigure}[b]{.185\textwidth}
        \centering
        \includegraphics[width=.8\textwidth]{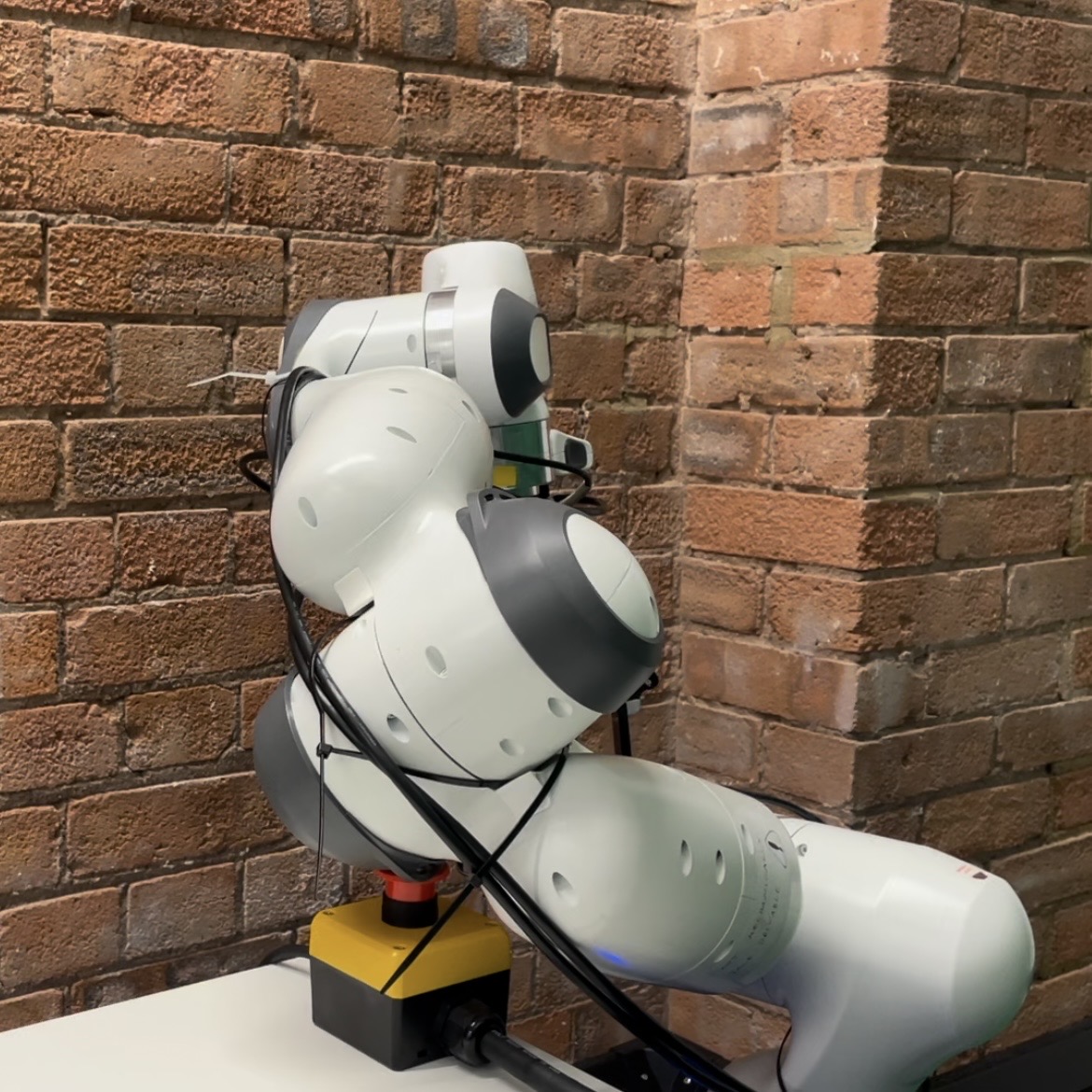}
        \caption{Push Button Elbow}
    \end{subfigure}
    \captionof{figure}{Pictures representative of the 5 real-world robotic tasks solved with imitation learning.  
    }
    \label{fig:real_setup}
\end{minipage}
\end{figure*}

\textbf{Full-body tasks.} Building on top of RLBench \cite{James2019RLBench}, we evaluate our approach on a set of 6 tasks where full control of the robot joint configuration is required to carry out the task successfully and consistently. Among these tasks, 4 tasks are completely new: `reach elbow pose' and `take bottle out fridge' are easy tasks that require very accurate control of the robot's configuration, while `slide cup obstacles' and `serve coffee obstacles' are more complex but only require obstacle avoidance. The remaining two tasks (`meat off grill' and `take cup out cabinet') are more complex tasks from the original RLBench tasks, which presents several variations where obstacle avoidance is necessary to consistently solve the task. 

Detailed final performance after 500k steps for each task is shown in Figure \ref{fig:elbow_tasks}. Overall, we observe that the performance obtained using ERJ is much higher than with task and joint spaces. ERA, instead, works well in simple tasks that require accurate configuration control, outperforming all other approaches, including ERJ, but loses in data efficiency when compared to ERJ and task in the other, more complex tasks.
Joint space actions allow solving simpler tasks, despite being less efficient than ERJ and ERA, but performs poorly elsewhere. \textcolor{new}{JAiLeR shows the potential to solve several of the tasks, especially when rewarding the learned controller with task rewards. However, the learned controller tends to slow down learning of the task and so the task performance remains low within 500k environment steps, compared to the other approaches.}
Task space control is sometimes able to solve more complex tasks that require obstacle avoidance, but ERJ performs more reliably, thanks to the more comprehensive control over the robot's configuration.  
To further highlight the learning efficiency of ERJ, performance over time across all the tasks is summarised in Figure \ref{fig:performance_comparison}. 

\textbf{Standard RLBench tasks.} We study the performance of the ER space in a set of standard RLBench tasks that have previously been used in several works~\cite{James2022ARM, liu2022auto, james2022coarse, zhao2022effectiveness, james2022te, adeniji2023language, seo2023multi}. These tasks do not require elbow positioning as there are no obstacles to avoid or precise poses to achieve. Thus, we use them to verify that the learning efficiency of task space is maintained for ER methods. 

In Figure \ref{fig:non_elbow_tasks}, we present performance over time. 
First, we note that, as expected, using the joint space (which is not aligned with the task space) leads to slower learning. \textcolor{new}{The performance of the JAiLeR-based baselines is close to the performance of Joint, mostly because learning the joint-level controller slows down the learning of the task policy.} Task space control efficiently and accurately solves all these tasks since no obstructions are present. The performance of ERJ is completely in line with task space, both in terms of final performance and learning efficiency, showing that the control over the additional joints doesn't slow down learning. ERA learns slower and we hypothesise this is due to the complexity of predicting valid \{end-effector, arm angle\} configurations. Our hypothesis is supported by the fact that ERA tends to produce more invalid actions, \textcolor{new}{as we analyse in more detail in Section \ref{subsec:ablation}.}

\subsection{Imitation Learning in Real World}

The use of ER is not limited to reinforcement learning; in this section we explore its use as an action space for real-world imitation learning \textcolor{new}{on a Franka Emika arm.}

\textcolor{new}{\textbf{Tasks description.} \textit{Reach \& Remove Cup.} This is a real-world adaptation of the `take cup out cabinet' task we adopted in simulation (see Figure \ref{fig:real_setup}). The cabinet has a simpler structure than that in simulation, but it is twice as deep (the real cabinet being $>0.3$ m deep); additionally, the narrow stem of the cup requires precise grasping in comparison with RLBench's simulated grasp mechanics. This task uses a wrist camera and the initial cup position is randomised within 3x3cm square.}
\textcolor{new}{We also divide 'Grasp \& Take Out Cup' into two sub-tasks:
(i) \textit{Reach Cup}: starting from the top of the cabinet, reaching the cup inside the cabinet and grasping it; 
(ii) \textit{Remove Cup}: starting from inside the cabinet, grasping and taking the cup out of the cabinet.
`Reach Cup' requires a correct change in elbow position to achieve success whilst `Remove Cup' can be achieved without changing the elbow position (since the sub-task is initialised inside the cupboard).}
\textcolor{new}{\textit{Retrieve Bear.} This task involves retrieving a bear from a cluttered cabinet top without colliding with any objects. Two wrist cameras and a shoulder camera are used and the initial bear position is randomised within a 10x6.5cm rectangle. The other objects on the cabinet top are also rotated ($\pm{15}^{\circ}$) and displaced (within a 3x3cm square) during training and testing.}
\textcolor{new}{\textit{Push Button Elbow.} A task where a button is pushing down with the elbow of the robot arm. This task uses a shoulder camera and the initial button position is randomised within a 10x1cm rectangle.}

\textcolor{new}{\textbf{Settings.} For each task we collect 40 demonstrations (32 for training and 8 for validation), which contain camera images and proprioceptive states (joint configuration and the pose of the end-effector) at each step. Solving these problems with imitation learning requires the action space to reproduce the full robot configuration seen in the demos due to either a confined workspace or non-gripper manipulation.} It is therefore expected that ER should be able to complete the tasks, whilst task space control success is not guaranteed.

For training the agent we use the same strategy as in ACT \cite{Zhao2023ACT}, \textcolor{new}{collecting demonstrations using a second 'leader' robot arm (another Franka Emika arm) to teleoperate the first 'follower' arm (which mirrors the joint positions),} and employing action chunking and 'temporal ensembling' over multi-step predictions. ACT has been shown to perform very robustly in Joint space, \textcolor{new}{so we report its performance on the tasks as an `oracle', but focus our comparison on the performance of task space and ERJ}. \textcolor{new}{For translating actions in the end-effector frame to joint actions, we adopt MoveIt\footnote{\url{https://moveit.ros.org/}} and PickNikRobitic\footnote{\url{https://github.com/PickNikRobotics/pick_ik}} with default parameters.} 

\textbf{Results.} In Table~\ref{tbl:real_results}, we summarise the results where each task was evaluated 10 times per method. \textcolor{new}{As expected, we observe that ERJ successfully solves these tasks; however, task space is not able to avoid obstacles. The only success observed for task space is `Remove Cup' when the arm starts inside the cabinet; otherwise, task space failed to reach the cup or bear objects due to colliding with obstacles. The sole failures for Joint and ERJ occurred when grasping the cup's thin stem failed.}

\begin{figure}[t]
    \centering
    \includegraphics[width=0.7\columnwidth]{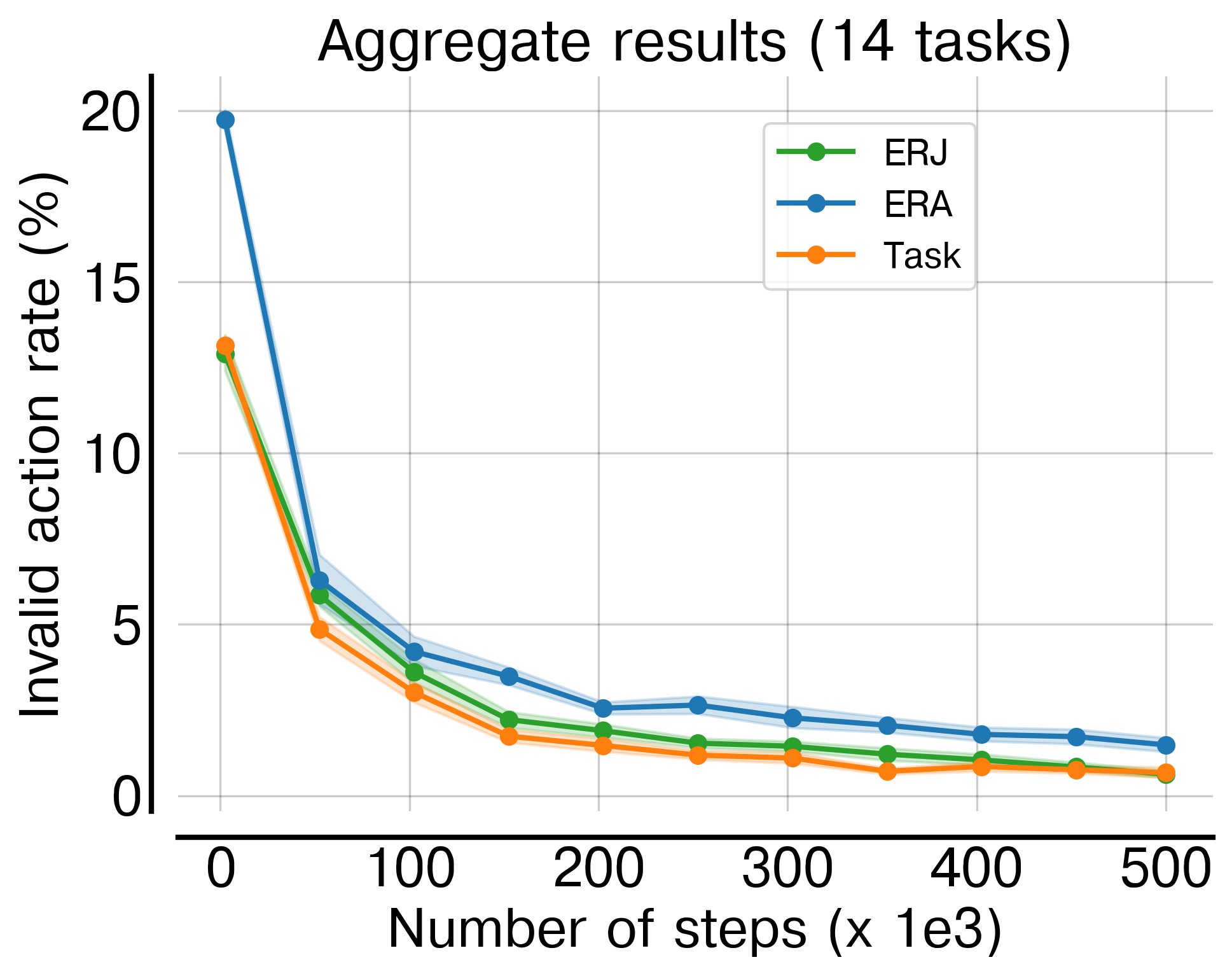}
    \caption{\textcolor{new}{Comparing IK-based action modalities in terms of invalid actions across all tasks (4 runs).}}
    \label{fig:invalid}
    \centering
    \includegraphics[width=0.65\linewidth]{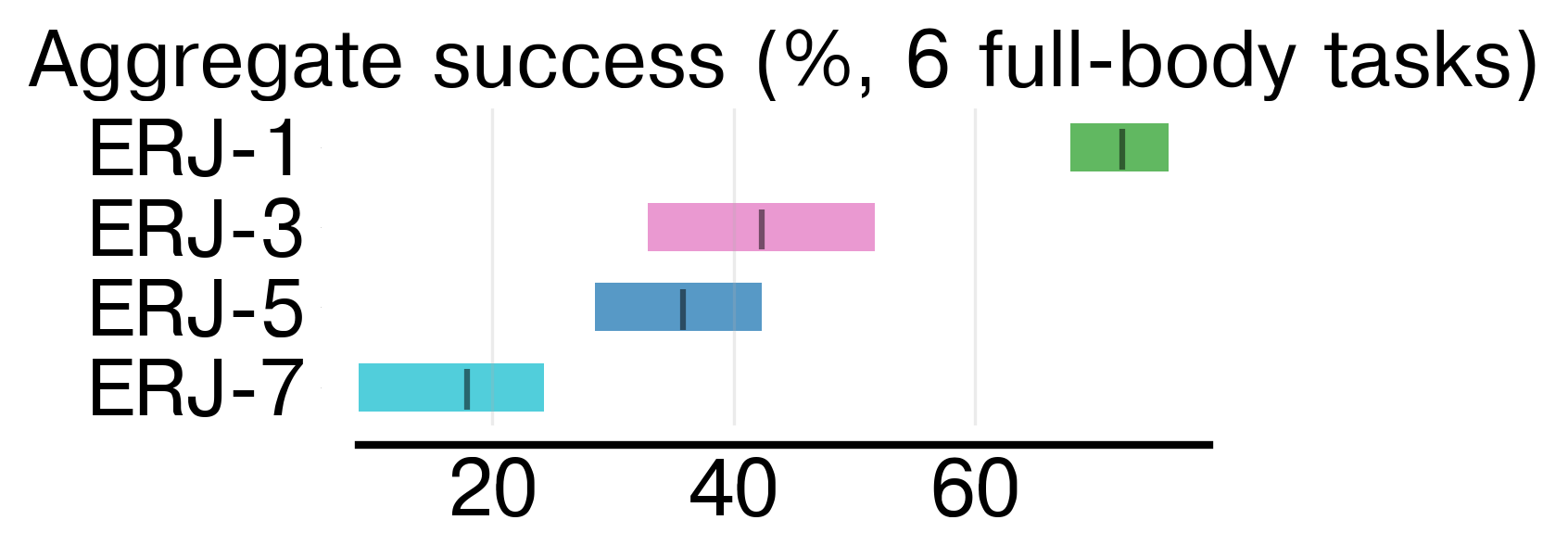}
    \caption{Comparing performance when controlling different joints for ERJ on the `Full-body' tasks \textcolor{new}{(3 runs)}.}
    \label{fig:joint_ablation}
\end{figure}%

\subsection{Additional studies}
\label{subsec:ablation}

\textcolor{new}{\textbf{Invalid actions in ERA.} When translating task space coordinates into joint commands using an IK solver, as done for task space, ERA and ERJ, the policy can provide unsolvable targets for the IK solver. Examples of these failure cases include: actions leading to unreachable areas outside of the robot's workspace, actions requiring impossible configurations due to the physical properties of the robot, or action displacements that are too large in magnitude, when using delta action spaces, as in our RL experiments.}

\textcolor{new}{In our settings, when an invalid action is attempted, i.e. the IK solver returns an error, the agent stops and is given a reward of zero, which discourages taking invalid actions (rewards are otherwise positive). Figure \ref{fig:invalid} shows the invalid action rate over time for different action modes. We observe that ERA’s invalid rate is initially much higher and the agent spends the initial steps learning how to provide valid configurations, with the consequence that the policy will learn to act using more conservative actions (small delta poses) in order to avoid invalid non-rewarding actions.} 

\textbf{Joint selection in ERJ.} In order to empirically verify the motivation for choosing the base joint (the first joint) to be controlled directly for ERJ, we perform an ablation study across the other joints. 
The results of training in simulation on the `Full-body tasks' are presented in Figure \ref{fig:joint_ablation}.

Joints numbered with even numbers are excluded from the Figure, as they all lead to performance collapse. On a 7 DoF robot, this is expected, as the robot is equipped with 4 twisting-joints (oddly numbered, e.g. first joint of a Panda) and 3 rotational-joints (evenly numbered, e.g. second joint of a Panda), thus the redundancy lies among the oddly numbered joints and constraining a rotational-joint can impair the robot's control. Overall, the results confirm the first joint to empirically be the most versatile choice, by a large margin. However, we note that in the `reach elbow pose' task, some joints actually allow faster learning than the first joint, showing that using the base joint is still an arbitrary choice and that there may be other tasks where direct control of a different joint would be more beneficial. We leave further studies on how to address this problem for future work. 

\begin{figure}[t]
    \centering
    \begin{subfigure}[t]{0.4\linewidth}
    \includegraphics[width=\textwidth]{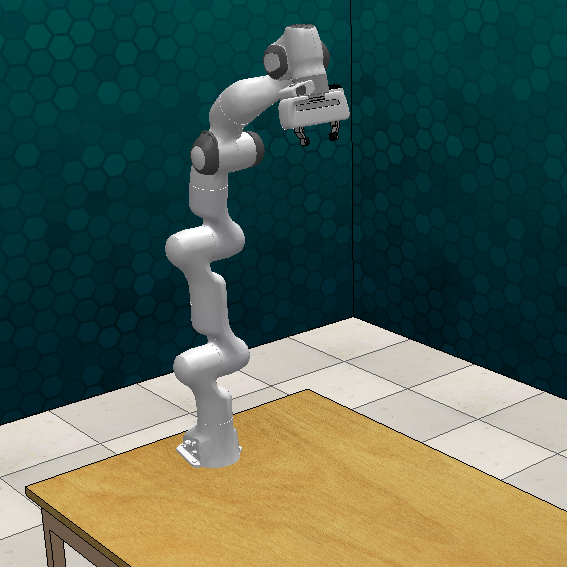}
    \end{subfigure}
    \hfill
    \begin{subfigure}[t]{0.56\linewidth}
    \includegraphics[width=\textwidth]{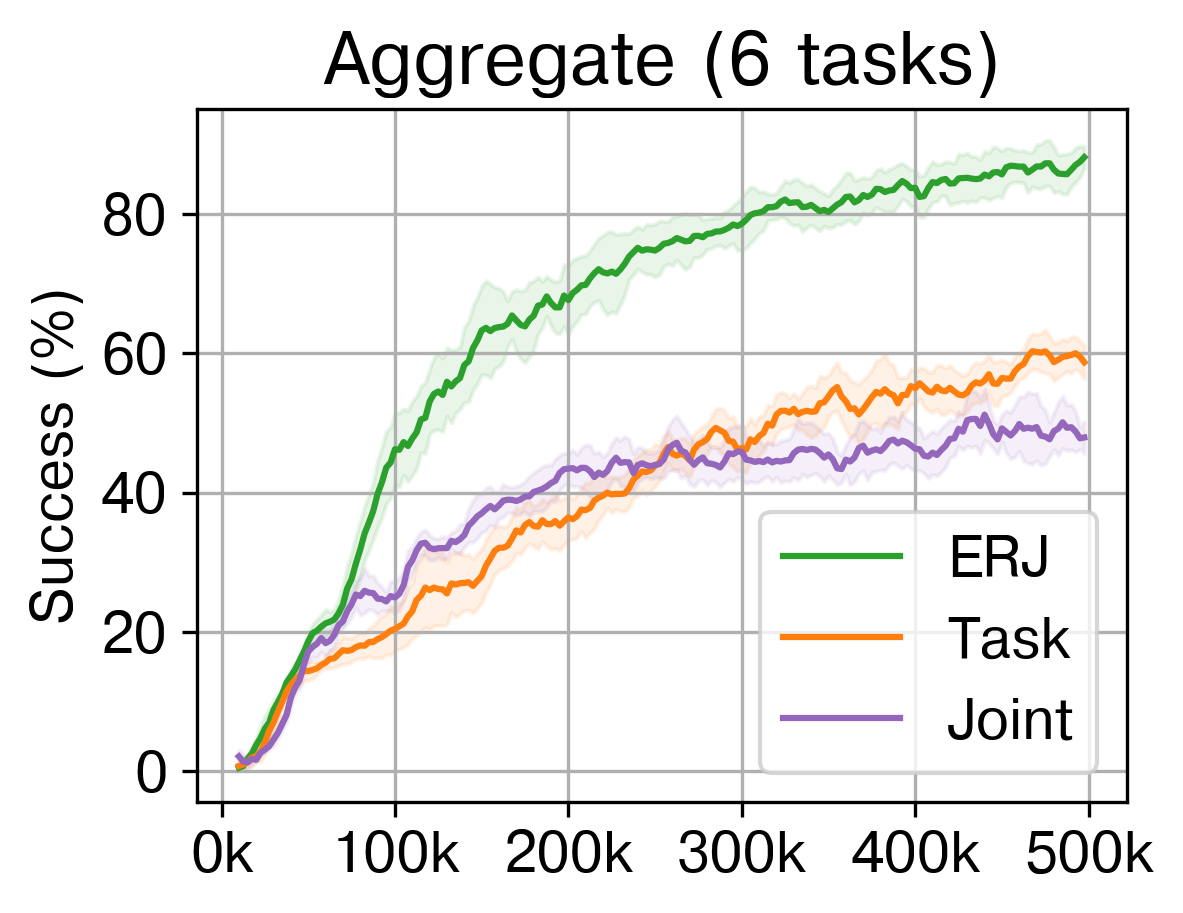}
    \end{subfigure}
    \hfill
    \begin{subfigure}[t]{\linewidth}
    \includegraphics[width=0.9\textwidth]{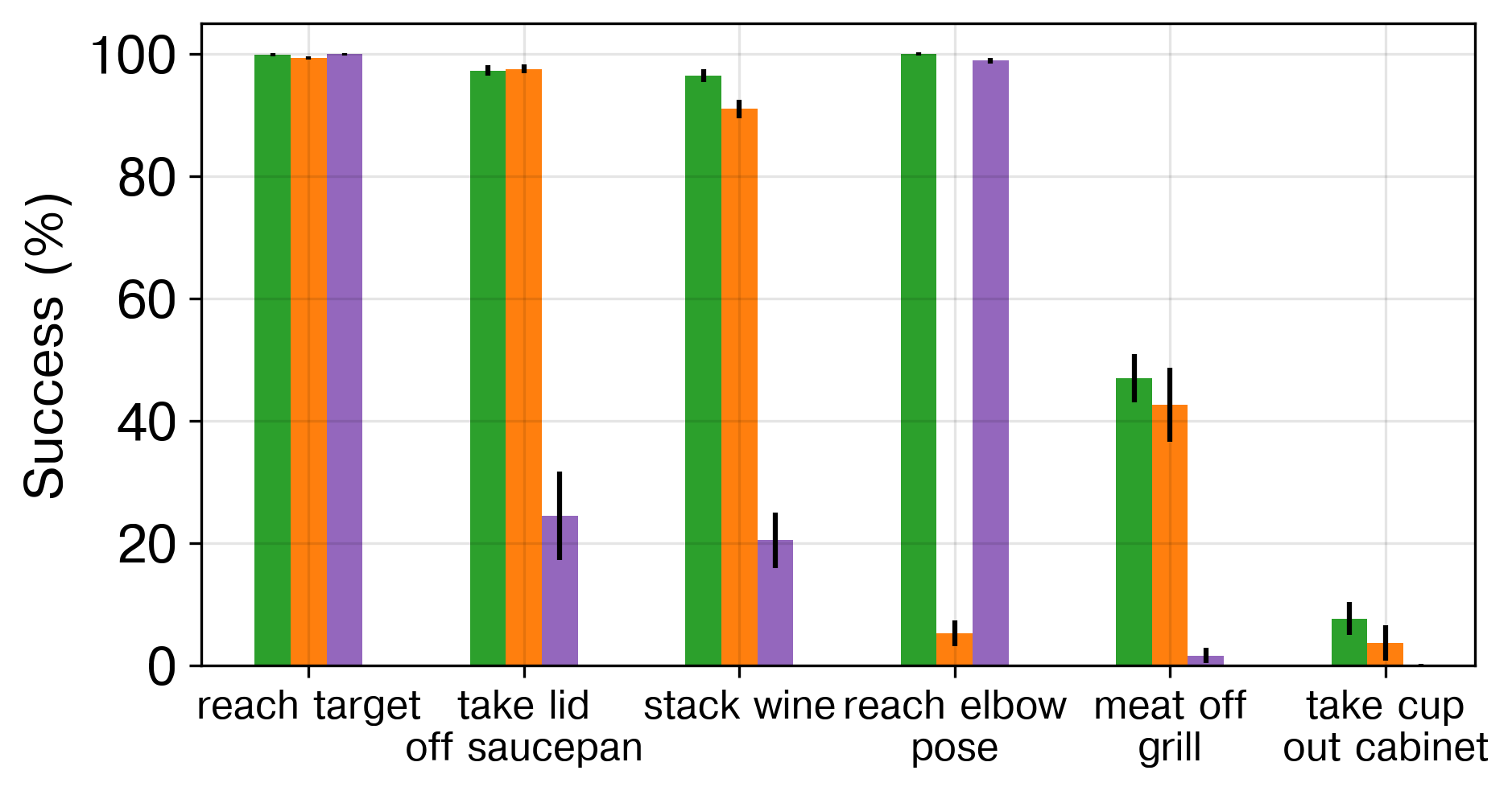}
    \end{subfigure}
    \caption{\textbf{Higher DoFs control} Performance when controlling a more complex robot (top left) with 8 DoF \textcolor{new}{(3 runs)}.}
    \label{fig:8dofs}
\end{figure}

\textbf{Higher DoFs control.} 
Given the lack of availability of robot models with more than 7 joints to control, in order to empirically verify the scalability of ERJ to robot arms with more than one redundant joint, we designed a custom 8 DoFs Panda arm. The arm has been modified to have two additional joints at the base of the robot, following the same structure of the original first two joints of the robot. The result is a Panda with 9 joints, of which we control 8 joints for simplicity, thus fixing the (new) base joint. In this setup, there are two redundant joints to control, one twisting-joint and one rotational-joint. For joint space, the setup requires controlling 8 joints. For task space, the setup requires adapting the IK computation to work with 8 joints. For ERJ, we separately control two joints (the first two controllable joints from the base) and, as done for 7 DoFs robots, we compute the IK on the remaining six joints (no redundancies).

\textcolor{new}{In Figure \ref{fig:8dofs}, we show the robot and compare both final and over-time performance for six tasks: three standard tasks, `reach target', `take lid off saucepan' and `stack wine', and three full-body tasks, `reach elbow pose', `meat off grill' and `take cup out cabinet'. We limited the evaluation to simple obstacle avoidance and unconstrained environment settings, otherwise switching to a new arm would require redesigning the tasks' environments to ensure compliance and solvability.  We note that the effects observed on the 7 DoFs Panda hold in this setup, with ERJ showing greater learning efficiency than Joint space control in the standard tasks, and being more performant than task space control in the fully-body tasks, overall obtaining the highest performance. }

\section{Conclusion}

We presented ER, a new action space formulation that addresses both precise control and efficient robot learning for overactuated arms, overcoming the shortcomings of the previous most common joint and task spaces. By putting emphasis on the issues which arise when not controlling the entire arm configuration, we show the importance of switching to new action representations as robot learning advances and moves on to more complex tasks, requiring precision and the ability to avoid obstacles. Our ERJ solution demonstrates ease of implementation and has exhibited consistent success in diverse settings, spanning both simulation and real-world scenarios. This suggests its potential readiness to universally supplant both task and joint space methods in the field of robot learning. 


Despite its advantages, singularity points still occur with ERJ, as they are rarer than with task space but still present. ERA does not present such problems but has demonstrated less reliable performance. Further investigation into strategies for mitigating singularities within ERJ or enhancing the efficiency/validity of the ERA space would prove valuable. Moreover, we have only investigated two parameterisations of the redundancy for ER spaces (angle and joint); future work could investigate alternative ways of implementing ER.

\section*{{\small ACKNOWLEDGEMENT}}
Pietro Mazzaglia is funded by a Ph.D. grant of the Flanders Research Foundation (FWO).




\bibliographystyle{IEEEtran}
\bibliography{IEEEabrv,references}

\addtolength{\textheight}{-12cm}   

\end{document}